\documentclass[10pt,twocolumn,letterpaper]{article}

\usepackage{xcolor}
\usepackage{cvpr}
\usepackage{times}
\usepackage{epsfig}
\usepackage{graphicx}
\usepackage{amsmath}
\usepackage{amssymb}
\usepackage{multirow}
\usepackage[vlined,boxed,commentsnumbered,ruled,linesnumbered]{algorithm2e}
\usepackage[pagebackref=true,breaklinks=true,colorlinks,bookmarks=false]{hyperref}

\usepackage{amsmath,amsfonts,bm}

\def\eqref#1{equation~\ref{#1}}

\def\1{\bm{1}}

\def\vs{{\bm{s}}}

\DeclareMathAlphabet{\mathsfit}{\encodingdefault}{\sfdefault}{m}{sl}
\SetMathAlphabet{\mathsfit}{bold}{\encodingdefault}{\sfdefault}{bx}{n}

\DeclareMathOperator*{\argmax}{arg\,max}

\cvprfinalcopy 

\def\cvprPaperID{5265} 

\ifcvprfinal\fi
\begin{document}


\newcommand{\xpar}[1]{\noindent\textbf{#1}\ \ }
\newcommand{\vpar}[1]{\vspace{3mm}\noindent\textbf{#1}\ \ }

\newcommand{\sect}[1]{Section~\ref{#1}}
\newcommand{\sects}[1]{Sections~\ref{#1}}
\newcommand{\eqn}[1]{Equation~\ref{#1}}
\newcommand{\eqns}[1]{Equations~\ref{#1}}
\newcommand{\fig}[1]{Figure~\ref{#1}}
\newcommand{\figs}[1]{Figures~\ref{#1}}
\newcommand{\tab}[1]{Table~\ref{#1}}
\newcommand{\tabs}[1]{Tables~\ref{#1}}

\newcommand{\ignorethis}[1]{}
\newcommand{\norm}[1]{\lVert#1\rVert}
\newcommand{\fcseven}{$\mbox{fc}_7$}

\renewcommand*{\thefootnote}{\fnsymbol{footnote}}

\def\naive{na\"{\i}ve\xspace}
\def\Naive{Na\"{\i}ve\xspace}

\makeatletter
\DeclareRobustCommand\onedot{\futurelet\@let@token\@onedot}
\def\@onedot{\ifx\@let@token.\else.\null\fi\xspace}

\def\iid{\emph{i.i.d}\onedot}
\def\eg{\emph{e.g}\onedot} \def\Eg{\emph{E.g}\onedot}
\def\ie{\emph{i.e}\onedot} \def\Ie{\emph{I.e}\onedot}
\def\cf{\emph{c.f}\onedot} \def\Cf{\emph{C.f}\onedot}
\def\etc{\emph{etc}\onedot} \def\vs{\emph{vs}\onedot}
\def\wrt{w.r.t\onedot} \def\dof{d.o.f\onedot}
\def\etal{\emph{et al}\onedot}
\def\modelshort{APQ\xspace}
\makeatother

\newcommand{\myparagraph}[1]{\vspace{-12pt}\paragraph{#1}}
\newcommand{\myrebutpara}[1]{\vspace{-14.2pt}\paragraph{#1}}

\newcommand{\motherNet}{once-for-all network}
\newcommand{\motherNetCap}{Once-for-all Network}
\newcommand{\childNet}{sub-network}
\newcommand{\childNetCap}{Sub-network}

\definecolor{citecolor}{RGB}{34,139,34}

\definecolor{MyDarkBlue}{rgb}{0,0.08,1}
\definecolor{MyDarkGreen}{rgb}{0.02,0.6,0.02}
\definecolor{MyDarkRed}{rgb}{0.8,0.02,0.02}
\definecolor{MyDarkOrange}{rgb}{0.40,0.2,0.02}
\definecolor{MyPurple}{RGB}{111,0,255}
\definecolor{MyRed}{rgb}{1.0,0.0,0.0}
\definecolor{MyGold}{rgb}{0.75,0.6,0.12}
\definecolor{MyDarkgray}{rgb}{0.66, 0.66, 0.66}


\newcommand{\SH}[1]{\textcolor{MyDarkRed}{[Song: #1]}}
\newcommand{\han}[1]{{\bf \color{MyDarkBlue} [Han: #1]}}
\newcommand{\kuan}[1]{{\bf \color{MyDarkBlue} [Kuan: #1]}}
\newcommand{\tianzhe}[1]{{\bf \color{MyDarkGreen} [Tianzhe: #1]}}
\newcommand{\zhijian}[1]{\textcolor{MyDarkBlue}{[Zhijian: #1]}}
\newcommand{\ji}[1]{\textcolor{orange}{[Ji: #1]}}
\newcommand{\hanrui}[1]{\textcolor{MyDarkBlue}{Hanrui: #1}}

\title{APQ: Joint Search for Network Architecture, Pruning and Quantization Policy}

\author{
\hspace{-20pt}Tianzhe Wang$^{1,2}$ \qquad
Kuan Wang$^1$\qquad
Han Cai$^{1}$\qquad
Ji Lin$^1$\qquad
Zhijian Liu$^{1}$ \qquad
Song Han$^1$ \\
$^1$Massachusetts Institute of Technology \quad \quad $^2$Shanghai Jiao Tong University \\
}

\maketitle

\begin{abstract}
We present APQ for efficient deep learning inference on resource-constrained hardware. Unlike previous methods that separately search the neural architecture, pruning policy, and quantization policy, we optimize them in a joint manner. To deal with the larger design space it brings, a promising approach is to train a quantization-aware accuracy predictor to quickly get the accuracy of the quantized model and feed it to the search engine to select the best fit.
However, training this quantization-aware accuracy predictor requires collecting a large number of quantized $\langle$model, accuracy$\rangle$ pairs, which involves quantization-aware finetuning and thus is highly time-consuming. To tackle this challenge, we propose to transfer the knowledge from a full-precision (i.e., fp32) accuracy predictor to the quantization-aware (i.e., int8) accuracy predictor, which greatly improves the sample efficiency. Besides, collecting the dataset for the fp32 accuracy predictor only requires to evaluate neural networks without any training cost by sampling from a pretrained once-for-all \cite{cai2019once} network, which is highly efficient.  
Extensive experiments on ImageNet demonstrate the benefits of our joint optimization approach. With the same accuracy, APQ reduces the latency/energy by \textbf{2}$\times$/\textbf{1.3}$\times$ over MobileNetV2+HAQ~\cite{sandler2018mobilenetv2,wang2019haq}. Compared to the separate optimization approach (ProxylessNAS+AMC+HAQ~\cite{cai2019proxylessnas,he2018amc,wang2019haq}), APQ achieves \textbf{2.3}\% higher ImageNet accuracy while reducing orders of magnitude GPU hours and CO$_2$ emission, pushing the frontier for green AI that is environmental-friendly.
The \href{https://github.com/mit-han-lab/apq}{code} and \href{https://youtu.be/s5v23hTe60s}{video} are publicly available. 

\end{abstract}

\section{Introduction}

Deep learning has prevailed in many real-world applications like autonomous driving, robotics, and mobile VR/AR, while efficiency is the key to bridge research and deployment. Given a constrained resource budget on the target hardware (\eg, latency, model size, and energy consumption), it requires an careful design of network architecture to achieve the optimal performance within the constraint. 
Traditionally, the deployment of efficient deep learning can be split into model architecture design and model compression (pruning and quantization). Some existing works~\cite{han2016deep,han2019design} have shown that such a sequential pipeline can significantly reduce the cost of existing models. Nevertheless, careful hyper-parameter tuning is required to obtain optimal performance~\cite{he2018amc}. The number of hyper-parameters grows exponentially when we consider the three stages in the pipeline together, which will soon exceed acceptable human labor bandwidth.

\begin{figure}[t]
    \centering
    \includegraphics[width=\linewidth]{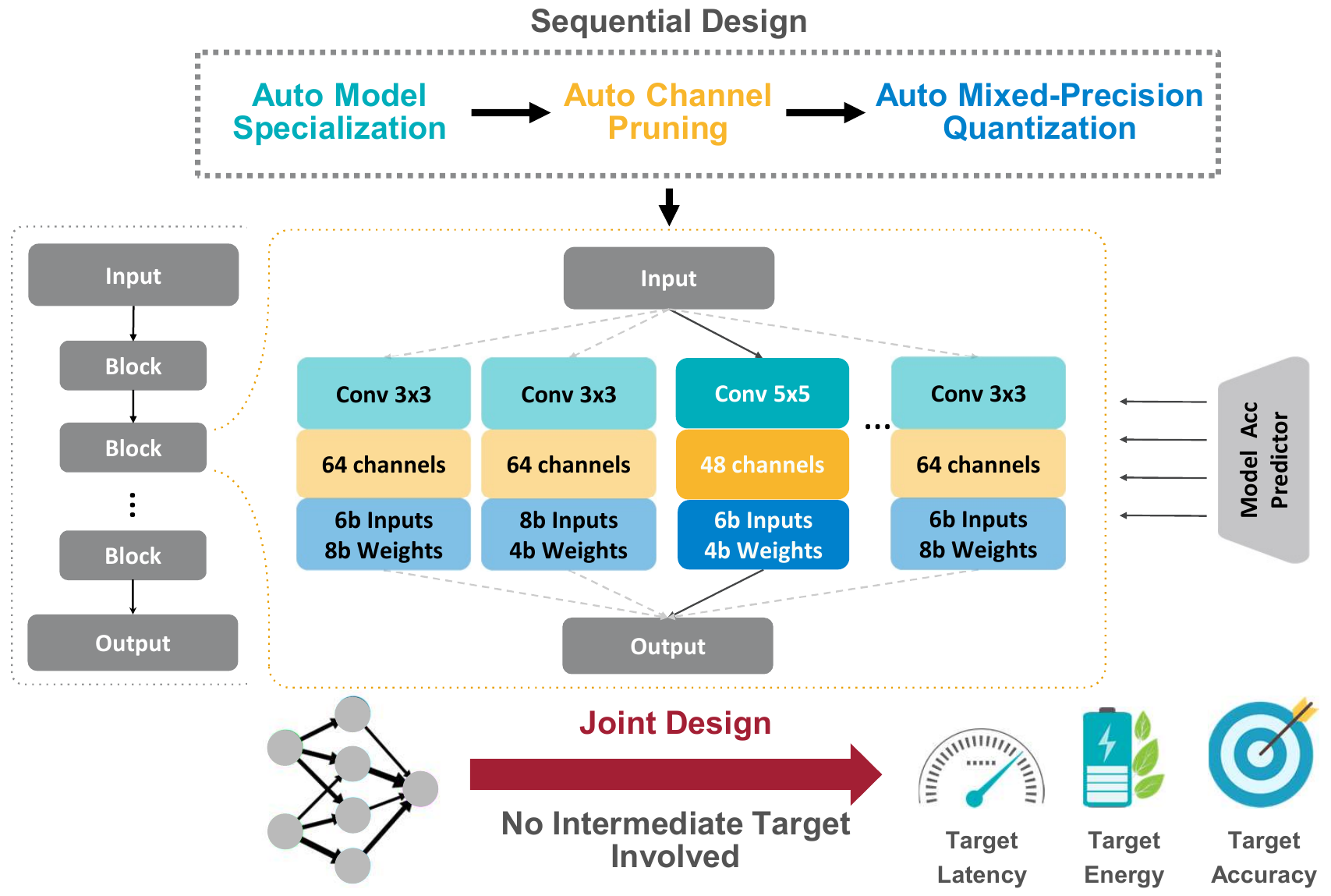}
    \caption{Comparison with sequential design and joint design. APQ combines the three optimization stages (architecture search, pruning, and quantization) into one stage, and jointly optimize for accuracy, latency and energy end-to-end.}
\label{fig:intro}
\end{figure}

To tackle the problem, recent works have applied AutoML techniques to automate the process. Researchers proposed Neural Architecture Search (NAS)~\cite{zoph2017neural, zoph2018learning, liu2017progressive, liu2018hierarchical, cai2018efficient, cai2018path, cai2019once, han2019design} to automate the model design, outperforming the human-designed models by a large margin. Using a similar technique, researchers adopt reinforcement learning to compress the model through automated pruning~\cite{he2018amc} and automated quantization~\cite{wang2019haq}. However, optimizing these three factors in separate stages will lead to sub-optimal results: \eg, the best network architecture for the full-precision model is not necessarily the optimal one after pruning and quantization. Besides, this three-step strategy also requires considerable search time and energy consumption~\cite{strubell2019energy}. Therefore, we need a solution to jointly optimize the deep learning model for a certain hardware platform.
\begin{table*}[!t]
\centering
\caption{Comparisons of architecture search approaches for efficient models (ProxylessNAS~\cite{cai2019proxylessnas}, SPOS: Single Path One-Shot~\cite{guo2019single}, ChamNet~\cite{dai2018chamnet}, AMC~\cite{he2018amc}, HAQ\cite{wang2019haq} and \modelshort (Ours). 
``No training during search'' means there is no need for re-training the sampled network candidate during search phase, and this is accomplished by \textit{once-for-all network} \cite{cai2019once}.
``No evaluation during search'' means that we do not have to evaluate sampled network candidate on validation dataset during search phase, and this is achieved by \textit{quantization-aware accuracy predictor} in \modelshort.
In a nutshell, \modelshort searches mixed-precision architecture without extra interaction (training or evaluation) with target dataset, which guarantees the low cost in search phase.
}
\vspace{5pt}
\begin{tabular}{lccccccc}
\hline
& ProxylessNAS & ChamNet & SPOS & AMC & HAQ & \textbf{\modelshort} \\
\hline
Hardware-aware      & \checkmark &  \checkmark  & \checkmark & \checkmark & \checkmark& \checkmark \\
No training during search      &   &  \checkmark  &  \checkmark & & & \checkmark \\
No evaluation during search     &   &    \checkmark & & & & \checkmark \\
Channel pruning   &   &    & &\checkmark &  & \checkmark \\
Mixed-precision quantization              &   &    & \checkmark  & & \checkmark & \checkmark \\
\hline
\end{tabular}
\label{tab:comparison_with_other_methods}
\end{table*}

\begin{figure}[ht]
    \centering
    \includegraphics[width=1\linewidth]{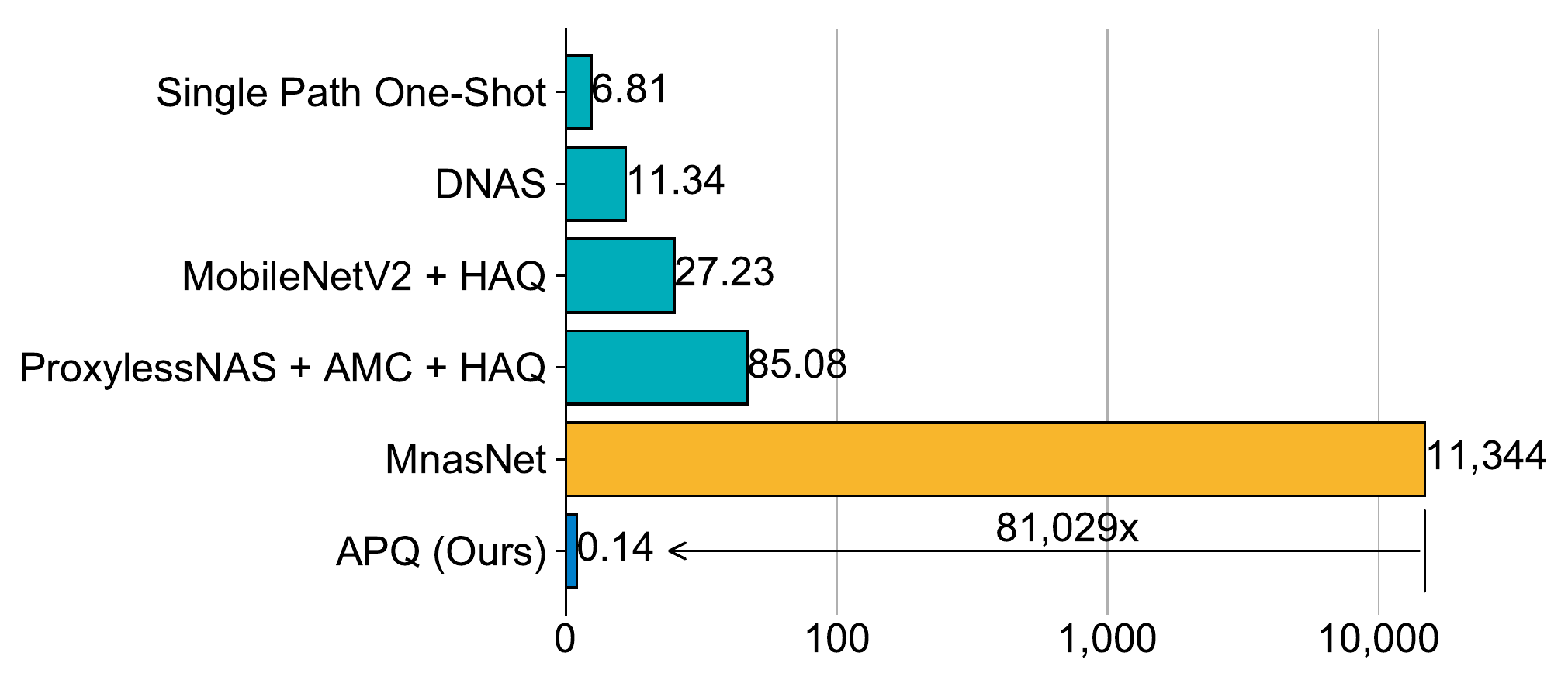}
    \caption{The illustration of marginal search cost for an upcoming scenario measured in pounds of CO$_2$ emission. Simply extending existing methods could still cost a considerable CO$_2$ emission which is not environmental-friendly.}
\label{fig:search_bar}
\end{figure}

Directly extending existing AutoML techniques to the joint model optimization setting can be problematic. Firstly, the joint search space is much larger (multiplicative) compared to the stage-wise search, making the search difficult.
Pruning and quantization usually requires time-consuming fine-tuning process to restore accuracy~\cite{wang2019haq, yang2018netadapt}, which dramatically increases the search cost. 
As shown in Fig.~\ref{fig:search_bar}, searching for each deployment (ProxylessNAS+AMC+HAQ) will lead to a considerable CO$_2$ emission, which can exacerbate the greenhouse effect and seriously deteriorate the environment. 
Moreover, 
each step has its own optimization objective (\eg, accuracy, latency, energy); the final policy of the pipeline always turns out to be sub-optimal.

To this end, we propose~\textit{\modelshort}, a joint design method to enable end-to-end search of model \underline{A}rchitecture, \underline{P}runing, and \underline{Q}uantization policy with light cost. The core idea of APQ is to use a \emph{quantization-aware accuracy predictor} to accelerate the search process. The predictor takes the model architecture and the quantization scheme as input, and can quickly predicts its accuracy. Instead of fine-tuning the pruned and quantized network to get the accuracy, we use the estimated accuracy generated by the predictor, which can be obtained with negligible cost (since the predictor requires only a few FC layers). 

However, training an accurate predictor is challenging: it requires a lot of (quantized model, quantized accuracy) data points to train an accurate predictor. Collecting each of the data points could be quite expensive: 1. we need to train the network to get the initial fp32 weights, 2. and further fine-tuning to get the quantized int8 weights to evaluate the accuracy. Both stages are quite expensive, requiring hundreds of GPU hours. 

Luckily, inspired by the weight sharing mechanism in recent one-shot neural architecture search methods~\cite{guo2019single, cai2019once}, we reduce the cost of stage 1 by training a super network that contains all the sub-networks in the search space through weight sharing, and directly evaluate the sub-network accuracy without further fine-tuning. As shown in~\cite{cai2019once}, it is possible to train a ``once-for-all'' super network that can support all the sub-networks while achieving on-par or even higher accuracy compared to training from scratch. In this way, we only need to evaluate the sub-network instead of training to get (fp32 model, fp32 accuracy) data points, which requires orders of magnitude smaller computation. 

Reducing the cost of stage 2 is more challenging. Typically, direct low-bit quantization without fine-tuning usually leads to near-zero accuracy. Therefore, fine-tuning is still needed to collect (quantized model,  quantized accuracy) data points. To reduce the cost of stage 2, we propose \emph{predictor-transfer technique}. Instead of collecting a lot of expensive (quantized model, quantized accuracy) data points to directly train the quantization-aware predictor, we first train a fp32 model accuracy predictor using the cheap (fp32 model, fp32 accuracy) data points collected with the weight-sharing once-for-all network  (evaluation only, no training required), and then transfer the predictor to the quantized model domain by fine-tuning it on a small number of expensive (quantized model, quantized accuracy) data points. The transfer technique dramatically improves the sample efficiency on the quantized network domain and reduces the overall cost to train the predictor.

After training this quantization-aware predictor $P(\text{arch, prune, quantization})$, the architecture search becomes ultra-fast by using the predictor. With the above design, we are able to efficiently perform a joint search over model architecture, channel number, and mixed-precision quantization. The predictor can also be used for new hardware and deployment scenarios.

Extensive experiments show the superiority of APQ. APQ achieves \textbf{8}$\times$ BitOps reduction than an 8-bit ResNet while having higher accuracy; APQ can not only optimize latency and accuracy, but also energy. We obtain the same accuracy as MobileNetV2+HAQ, and achieve \textbf{2}$\times$/\textbf{1.3}$\times$ latency/energy saving; APQ outperforms separate sequential optimizations using ProxylessNAS+AMC+HAQ by \textbf{2.3}\% accuracy under same latency constraints, while reducing \textbf{600}$\times$ GPU hours and CO$_2$ emission, which efficiently search an efficient model, pushing the frontier for green AI that is environmental-friendly.

The contributions of this paper are:
\begin{itemize}
    \item We propose a methodology to jointly perform NAS-pruning-quantization,  unifying the conventionally separated stages into an integrated solution.
    \item We propose a \textit{predictor-transfer} method to tackle the high cost of the quantization-aware accuracy predictor's dataset collection $\langle$NN architecture, quantization policy, quantized accuracy$\rangle$. 
    \item We achieve significant speedup to search optimal network architecture with quantization policy via this joint optimization, and enable automatic model adjustment in diverse deployment scenarios.
\end{itemize}

\section{Background and Outline}

\begin{figure*}[t]
    \centering
    \includegraphics[width=0.8\linewidth]{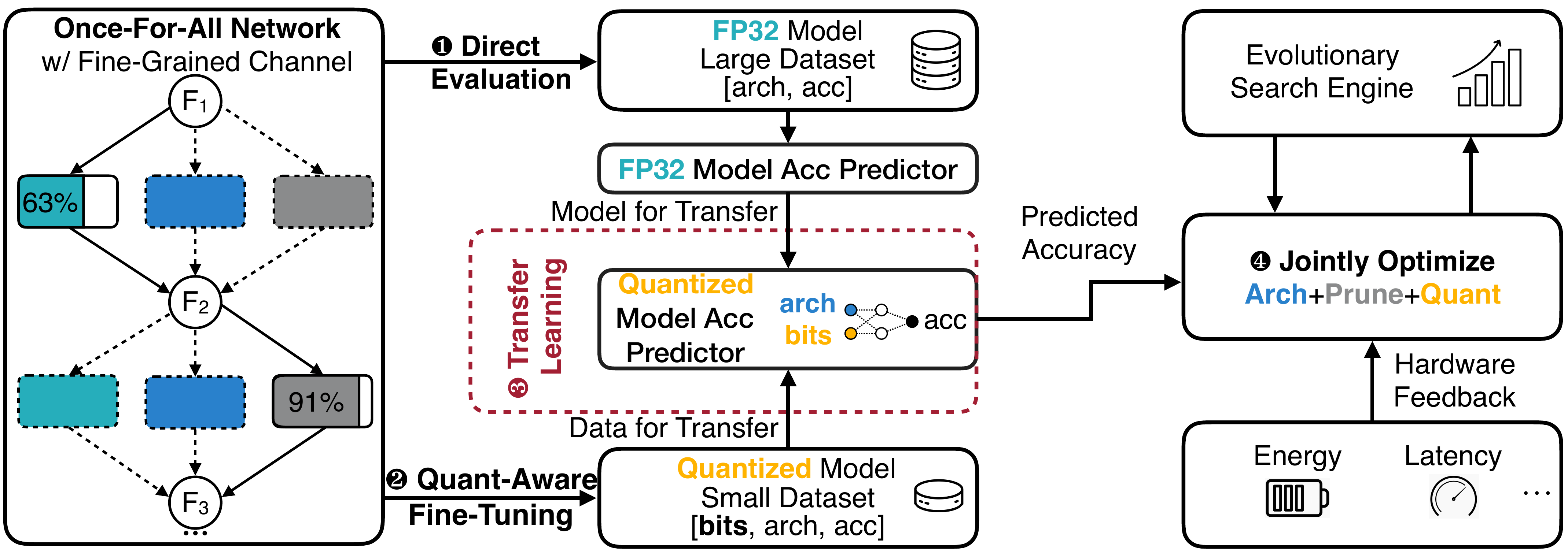}
    \caption{An overview of APQ's joint design methodology. The serial number represents the order of the steps. We first train an accuracy predictor for the full precision NN, then incrementally train an accuracy predictor for the quantized NN (predictor-transfer). Finally, evolutionary search is performed to find the specialized NN architecture with quantization policy that fits hardware constraints.}
\label{fig:overview}
\end{figure*}

Researchers have proposed various methods to accelerate the model inference, including architecture design~\cite{howard2017mobilenets,sandler2018mobilenetv2},  network pruning~\cite{han2015learning, liu2017learning} and network quantization~\cite{han2016deep}.

\myparagraph{Neural Architecture Search.}

Tracing back to the development of NAS, one can see the reduction in the search time. Former NAS~\cite{zoph2018learning, real2018regularized} use an RL agent to determine the cell-wise architecture. To efficiently search for the architecture, many later works viewed architecture searching as a pathfinding problem~\cite{liu2019darts, cai2019proxylessnas}, it cuts down the search time by jointly training rather than iteratively training from scratch. Inspired by the path structure, some one-shot methods~\cite{guo2019single} have been proposed to further leverage the network's weights in training time and begin to handle the mixed-precision case for efficient deployment. Another line of works tries to grasp the information by a performance predictor \cite{luo2018neural, dai2018chamnet}, which reduces the frequent evaluation for the target dataset when searching for optimal.

\myparagraph{Pruning.}
Extensive works show the progress achieved in pruning: in the early time, researchers proposed fine-grained pruning~\cite{han2015learning,han2016deep} by cutting off the connections (\ie, elements) within the weight matrix. However, such kind of method is not friendly to the CPU and GPU, and requires dedicated hardware\cite{pal2018outerspace, sparch} to support sparse matrix multiplication, which is highly demanding to design \cite{learncircuits, learncircuits2, mao2019park}. Later, some researchers proposed channel-level pruning~\cite{he2017channel,liu2017learning, Lin:2017ww, molchanov2016pruning, anwar2016compact, hu2016network, Polyak2015channel} by pruning the entire convolution channel based on some importance score (\eg, L1-norm) to enable acceleration on general-purpose hardware. However, both fine-grained pruning and channel-level pruning introduces an enormous search space as different layer has different sensitivities (\eg, the first convolution layer is very sensitive to be pruned as it extracts important low-level features; while the last layer can be easily pruned as it's very redundant). To this end, recent researches leverage the AutoML techniques~\cite{he2018amc, yang2018netadapt} to automate this exploration process and surpass the human design.
\myparagraph{Quantization.}
Quantization is a necessary technique to deploy the models on hardware platforms like FPGAs and mobile phones. \cite{han2016deep} quantized the network weights to reduce the model size by grouping the weights using k-means. \cite{Courbariaux:2016tm} binarized the network weights into $\{-1, +1\}$; \cite{Zhou:2016wh} quantized the network using one bit for weights and two bits for activation; \cite{Rastegari:2016tn} binarized each convolution filter into $\{-w, +w\}$; \cite{zhu2016trained} mapped the network weights into $\{-w_\text{N}, 0, +w_\text{P}\}$ using two bits with a trainable range; \cite{zhou2018explicit} explicitly regularized the loss perturbation and weight approximation error in a incremental way to quantize the network using binary or ternary weights. \cite{Jacob:2018ur} used 8-bit integers for both weights and activation for deployment on mobile devices. 
Some existing works explored the relationship between quantization and network architecture. HAQ~\cite{wang2019haq} proposed to leverage AutoML to determine the bit-width for a mixed-precision quantized model. A better trade-off can be achieved when different layers are quantized with different bits, showing the strong correlation between network architecture and quantization.

\myparagraph{Multi-Stage Optimization.}
Above methods are orthogonal to each other and a straightforward combination approach is to apply them sequentially in multiple stages \ie NAS+Pruning+Quantization:
\begin{itemize}
    \item In the first stage, we can search the neural network architecture with the best accuracy on the target dataset~\cite{tan2018mnasnet,cai2019proxylessnas,wu2018fbnet}:
    \begin{equation}
        \mathcal{A}^{*}, w^{*} = \argmax_{\mathcal{A}, w} ACC_{val} \big(\mathcal{A}, w \big).
    \end{equation}
    where $ACC_{val} \big(\mathcal{A}, w \big)$ denotes the validation accuracy given a model with architecture $\mathcal{A}$ and weight $w$.
    \item In the second stage, we can prune the channels in the model automatically~\cite{he2018amc}:
    \begin{equation}
        P^{*} = \argmax_{P} ACC_{val} \big(P(\mathcal{A}^{*},w^{*}) \big).
    \end{equation}
    where $P(\mathcal{A}, w)$ outputs a pair $(\mathcal{A}', w')$ denoting the model architecture and fine-tuned weight after applying certain pruning policy $P$.
    \item In the third stage, we can quantize the model to mixed-precision \cite{wang2019haq}:
    \begin{align}
        \mathcal Q^{*} = \argmax_{Q} ACC_{val} \big(Q({P}^{*}(\mathcal{A}^{*},w^{*})) \big)
    \end{align}
    where $Q(\mathcal{A}, w)$ outputs a pair $(\mathcal{A}', w')$ denoting the model architecture and fine-tuned weight after applying certain quantization policy $Q$.
\end{itemize}
However, this \emph{separation} usually leads to a sub-optimal solution: \eg, the best neural architecture for the floating-point model may not be optimal for the quantized model. Moreover, frequent evaluations on the target dataset make such kind of methods time-costly: \eg, a typical pipeline as above can take about 300 GPU hours, making it hard for researchers with limited computation resources to do automatic design.
\myparagraph{Joint Optimization.}

Instead of optimizing NAS, pruning and quantization independently, joint optimization aims to find a balance among these configurations and search for the optimal strategy.
To this end, the joint optimization objective can be formalized into:
\begin{equation}\small
\label{eq:objective}
    \mathcal{A}^{*},w^{*},P^{*},Q^{*}= \argmax_{\mathcal{A},w,P,Q} ACC_{val} \big( Q(P(\mathcal{A}, w)) \big),
\end{equation}
However, the search space of this new objective is tripled as original one, so it becomes challenging to perform joint optimization.
We endeavor to unify NAS, pruning and quantization as joint optimization. The outline is:
1. Train a once-for-all network that covers a large search space and every sub-network can be directly extracted without re-training. 
2. Build a quantization-aware accuracy predictor to predict quantized accuracy given a sub-network and quantization policy. 
3. Construct a latency/energy lookup table and do resource constrained evolution search. 
Thereby, this optimization problem can be tackled jointly.

\section{Joint Design Methodoloy}

The overall framework of our joint design is shown in Figure~\ref{fig:overview}. It consists of a highly flexible once-for-all network with fine-grained channels, an accuracy predictor, and evolution search to jointly optimize the architecture, pruning, and quantization.

\subsection{Once-For-All Network with Fine-grained Channel Pruning}

Neural architecture search aims to find a good sub-network from a large search space. Traditionally, each sampled network is trained to obtain the actual accuracy~\cite{zoph2017neural}, which is time-consuming. Recent one-shot based NAS~\cite{guo2019single} first trains a large, multi-branch network. At each time, a sub-network is extracted from the large network to directly evaluate the approximated accuracy. Such a large network is called \emph{once-for-all network}. Since the choice of different layers in a deep neural network is largely independent, a popular way is to design multiple choices (\eg, kernel size, expansion ratios) for each layer.

In this paper, we used MobileNetV2 as the backbone to build a once-for-all network that supports different kernel sizes (\ie 3, 5, 7) and channel number (\ie 4$\times$$\mathcal{B}$ to 6$\times$$\mathcal{B}$, 8 as interval, $\mathcal{B}$ is the base channel number in that block) in block level, and different depths (\ie 2, 3, 4) in stage level. The combined search space contains more than $10^{35}$ sub-networks, which is large enough to perform the search on the top of it.

\myparagraph{Properties of the Once-For-All Network.}
To ensure efficient architecture search, we find that the once-for-all network needs to satisfy the following properties:
(1) For every extracted sub-network, the performance could be directly evaluated without re-training, so that the cost of training only needs to be paid once. 
(2) Support an extremely large and fine-grained search space to support channel number search. As we hope to incorporate pruning policy into architecture space, the once-for-all network not only needs to support different operators, but also fine-grained channel numbers (8 as interval). 
Thereby, the new space is significantly enlarged (nearly quadratic from $10^{19}$ to $10^{35}$). 

However, it is hard to achieve the two goals at the same time due to the nature of once-for-all network training: it is generally believed that if the search space gets too large (\eg, supporting fine-grained channel numbers), the accuracy approximation would be inaccurate~\cite{liu2019metapruning}. A large search space will result in high variance when training the once-for-all network.
To address the issue, 
we adopt progressive shrinking (PS) algorithm~\cite{cai2019once} to train the once-for-all network. Specifically, we first train a full sub-network with the largest kernel sizes, channel numbers and depths in the once-for-all network, and use it as a teacher to progressively distill the smaller sub-networks sampled from the once-for-all network. During distillation, the trained sub-networks still update the weights to prevent accuracy loss. The PS algorithm effectively reduces the variance during once-for-all network training. By doing so, we can assure that the extracted sub-network from the once-for-all network preserves competitive accuracy without re-training.

\begin{figure}[t]
    \centering
    \includegraphics[width=0.75\linewidth]{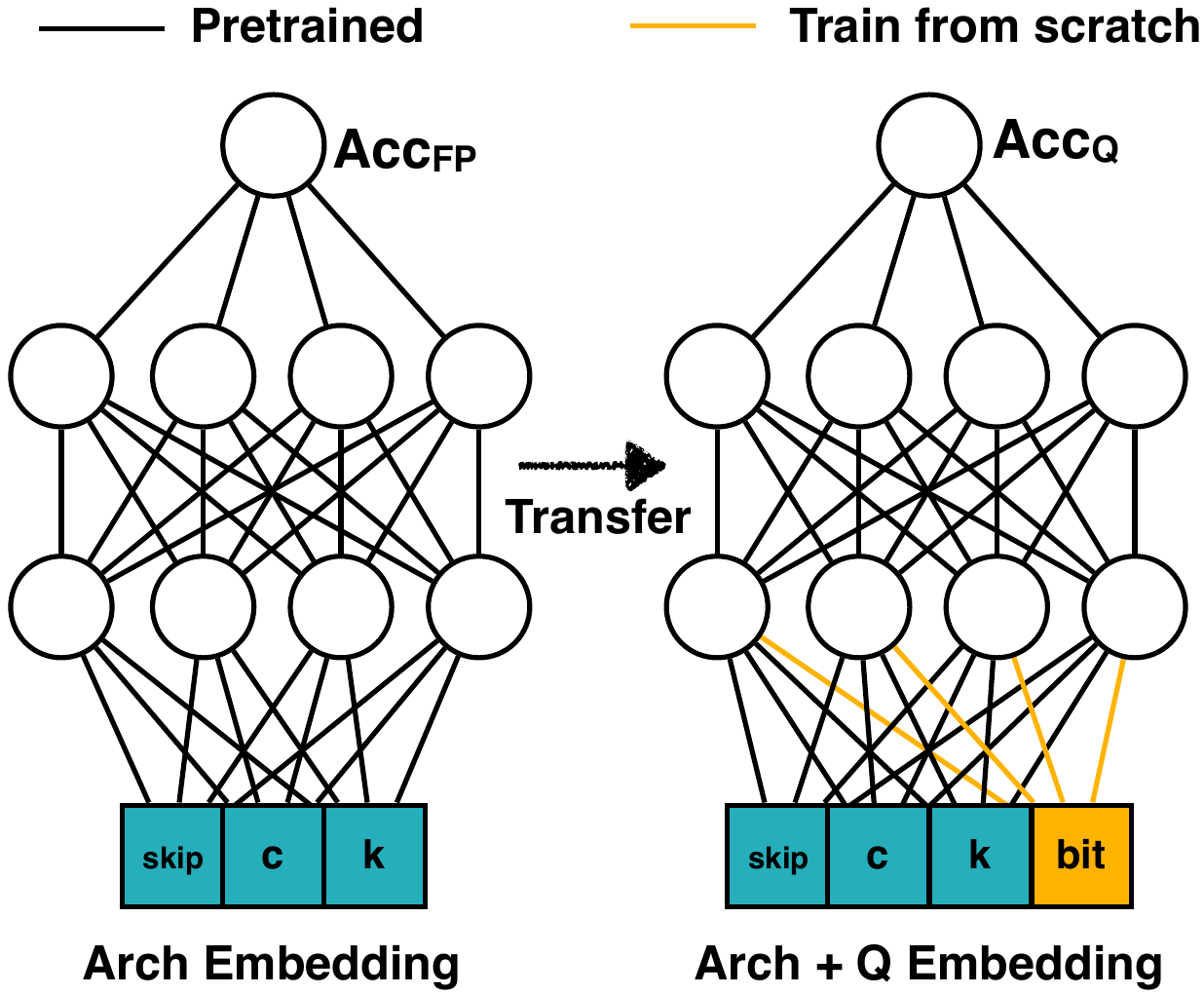}
    \caption{Predictor-transfer technique. We start from a pre-trained full-precision predictor and add another input head (yellow square at bottom right) denoting quantization policy. Then fine-tune the quantization-aware accuracy predictor.}
\label{fig:transfer}
\end{figure}

\subsection{Quantization-Aware Accuracy Predictor}
To reduce the cost for designs in various deployment scenarios, we propose to build a quantization-aware accuracy predictor $P$, which predicts the accuracy of the mixed-precision (MP) model based on architecture configurations and quantization policies. During search, we used the predicted accuracy $\overline{acc} = P(\text{arch, prune, quantize})$ instead of the measured accuracy. The input to the predictor $P$ is the encoding of the network architecture, the pruning strategy, and the quantization policy.

\begin{algorithm}[tb!] 
	\caption{\textbf{APQ framework}}
	\label{alg:tree-decay}
	\begin{small}
	\KwIn{Pretrained once-for-all network $\mathcal{S}$, evolution round $iterMax$, population size $N$, mutation rate $prob$, architecture constraints $C$.}
	Use $\mathcal{S}$ to generate FP32 model dataset $\mathcal{D}_{FP}$ $\langle$arch, acc$\rangle$ and quantized model dataset $\mathcal{D}_{MP}$ $\langle$quantization policy, arch, acc$\rangle$.
	
    Use $\mathcal{D}_{FP}$ to train a full precision (FP) accuracy predictor $\mathcal{M}_{FP}$.
    
	Use $\mathcal{D}_{MP}$ and $\mathcal{M}_{FP}$ (pretrained weight to transfer) to train a mixed precision (MP) accuracy predictor $\mathcal{M}_{MP}$.
	
	Randomly generate initial population $\mathcal{P}$ $\langle$quantization policy, arch$\rangle$ with size $N$ satisfying $C$.
	
	\For{i = 1 \dots iterMax}
	{   
	    Use $\mathcal{M}_{MP}$ to predict accuracy for candidates in $\mathcal{P}$ and update $Top_{k}$ with the candidates having Top-k highest accuracy. \\
        $\mathcal{P}_{crossover} = Crossover(Top_{k}, N/2, C)$ \\
        $\mathcal{P}_{mutation} = Mutation(Top_{k}, N/2, prob, C)$ \\
        $\mathcal{P} = Top_k\cup \mathcal{P}_{crossover} \cup \mathcal{P}_{mutation}$ \\
	}
	\KwOut{Candidates with best accuracy $Top_k$.}
	\end{small}
\end{algorithm}

\myparagraph{Architecture and Quantization Policy Encoding.}
We encode the network architecture block by block: for each building block (\ie bottleneck residual block like MobileNetV2~\cite{sandler2018mobilenetv2}), we encode the kernel size, channel numbers, weight/activation bits for pointwise and depthwise convolutions into one-hot vectors, and concatenate these vectors together as the encoding of the block. For example, a block has 3 choices of kernel sizes (\eg 3,5,7) and 4 choices of channel numbers (\eg 16,24,32,40), if we choose kernel size=3 and channel numbers=32, then we get two vectors [1,0,0] and [0,0,1,0], and we concatenate them together and use [1,0,0,0,0,1,0] to represent this block's architecture. Likewise, we also use one-hot vectors to denote the choice of bitwidth for certain weights/activation of pointwise and depthwise layers, \eg suppose weight/activation bitwidth choices for pointwise/depthwise layer are 4 or 8, we use [1,0,0,1,0,1,1,0] to denote the choice (4,8,8,4) for quantization policy. If this block is skipped, we set all values of the vector to 0. We further concatenate the features of all blocks as the encoding of the whole network. Then for a 5-layer network, we can use a 75-dim (5$\times$(3+4+2$\times$4)=75) vector to represent such an encoding. In our setting, the choices of kernel sizes are [3,5,7], the choices of channel number depend on the base channel number for each block, and bitwidth choices are [4,6,8], there are 21 blocks in total to design.

\myparagraph{Accuracy Predictor.}
The predictor we use is a 3-layer feed-forward neural network with each embedding dim equaling to 400. As shown in the left of Figure \ref{fig:transfer}, the input of the predictor is the one-hot encoding described above and the output is the predicted accuracy. Different from existing methods~\cite{liu2019darts, cai2019proxylessnas, wu2018fbnet}, our predictor based method does not require frequent evaluation of architecture on the target dataset in the search phase. Once we have the predictor, we can integrate it with any search method (\eg reinforcement learning, evolution, bayesian optimization, etc.) to perform joint design over architecture-pruning-quantization at a negligible cost. However, the biggest challenge is how to collect the $\langle$architecture, quantization policy, accuracy$\rangle$ dataset to train the predictor for quantized models, which is due to: 1) collecting quantized model's accuracy is time-consuming: fine-tuning is required to recover the accuracy after quantization, which takes about 0.2 GPU hours per data point. 
In fact, we find that for training a good full precision accuracy predictor, 80k $\langle$NN architecture, ImageNet accuracy$\rangle$ data pairs would be enough. 
However, if we collect a quantized dataset with the same size as the full precision one, it can cost 16,000 GPU hours, which is far beyond affordable. 2) The quantization-aware accuracy predictor is harder to train than the traditional accuracy predictor on full-precision models: the architecture design and quantization policy affect network performance from two separate aspects, making it hard to model the mutual influence. Thus using the traditional way to train quantization-aware accuracy predictor can result in a significant performance drop (Table \ref{tab:comparison_with_nas}).

\myparagraph{Transfer Predictor to Quantized Models.} 
Collecting a quantized NN dataset for training the predictor is difficult (needs finetuning), but collecting a full-precision NN dataset is easy: we can directly pick sub-networks from the once-for-all network and measure its accuracy. 
We propose the \textit{predictor-transfer technique} to increase the sample efficiency and make up for the lack of data.
As the order of accuracy before and after quantization is usually preserved, we first pre-train the predictor on a large-scale dataset to predict the accuracy of full-precision models, then transfer to quantized models. The quantized accuracy dataset is much smaller and we only perform short-term fine-tuning.
As shown in Figure~\ref{fig:transfer}, we add the quantization bits (weights\& activation) of the current block into the input embedding to build the quantization-aware accuracy predictor.
We then further fine-tune the quantization-aware accuracy predictor using pre-trained FP predictor's weights as initialization. 
Since most of the weights are inherited from the full-precision predictor, the training requires much fewer data compared to training from scratch.

\subsection{Hardware-Aware Evolutionary Search}

As different hardware might have drastically different properties (\eg, cache size, level of parallelism), the optimal network architecture and quantization policy for one hardware are not necessarily the best for the other. Therefore, instead of relying on some indirect signals (\eg, BitOps), our optimization is directly based on the measured latency and energy on the target hardware.
\myparagraph{Measuring Latency and Energy.}

Evaluating each candidate policy on actual hardware can be very costly. Thanks to the sequential structure of the neural network, we can approximate the latency (or energy) of the model by summing up the latency (or energy) of each layer. We can first build a lookup table containing the latency and energy of each layer under different architecture configurations and bit-widths. Afterward, for any candidate policy, we can break it down and query the lookup table to directly calculate the latency (or energy) at negligible cost. In practice, we find that such practice can precisely approximate the actual inference cost.

\myparagraph{Resource-Constrained Evolution Search.}

We adopt the evolution-based architecture search~\cite{guo2019single} to explore the best resource-constrained model. Based on this, we further replace the evaluation process with our \emph{quantization-aware accuracy predictor} to estimate the performance of each candidate directly. The cost for each candidate can then be reduced from $N$ times of model inference to only \textbf{one time} of predictor inference (where $N$ is the size of the validation set). Furthermore, we can verify the resource constraints by our \emph{latency/energy lookup table} to avoid the direct interaction with the target hardware. Given a resource budget, we directly eliminate the candidates that exceed the constraints.

\section{Implementation Details}
\label{sec:detail}
\vspace{8pt}
\myparagraph{Data Preparation for Quantization-aware Accuracy Predictor.}
We generate two kinds of data (2,500 for each): 1. random sample both architecture and quantization policy; 2. random sample architecture, and sample 10 quantization policies for each architecture configuration. We mix the data for training the quantization-aware accuracy predictor, and use full-precision pretrained predictor's weights to transfer. The number of data to train a full precision predictor is 80,000. As such, our quantization accuracy predictor can have the ability to generalize among different \textit{architecture/quantization policy} pairs and learn the mutual relation between architecture and quantization policy. 

\myparagraph{Evolutionary Architecture Search.}
For evolutionary architecture search, we set the population size to be 100, and choose Top-25 candidates to produce the next generation (50 by mutation, 50 by crossover). Each population is a network architecture with a quantization policy, using the same encoding as a quantization-aware accuracy predictor. The mutation rate is 0.1 for each layer, which is the same as that in ~\cite{guo2019single}, and we randomly choose the new kernel size and channel number for mutation. For a crossover, each layer is randomly chosen from the layer configuration of its parents. We set max iterations to 500, and choose the best candidate among the final population.

\begin{table*}[!t]
\centering
\caption{Comparison with state-of-the-art efficient models for hardware with fixed quantization or mixed precision.
Our method cuts down the marginal search time by two-order of magnitudes while achieving better performance than others. The marginal CO$_2$ emission (lbs) and cloud compute cost (\$)~\cite{strubell2019energy} is negligible for search in a new scenario. Here marginal cost means the cost for searching in a new deployment scenario, we use $N$ to denote the number of up-coming deployment scenarios and we include the cost for training our once-for-all network in the "design cost". The listed "our models" are searched under different latency constraints for fair comparison.}
\vspace{5pt}
\small
\begin{tabular}{l | c | c | c | c | c | c | c }
\hline
\multirow{2}{*}{Model} & ImageNet & Latency  & Energy & BitOps & Design cost  & CO$_2$e & Cloud compute cost \\
& Top1 (\%) & (ms) & (mJ) & (G) & (GPU hours) & (marginal) & (marginal) \\
\hline
MobileNetV2 - 8bit & 71.8 & 9.10 & 12.46 & 19.2 & - & - & - \\
ProxylessNAS - 8bit& 74.2 & 13.14 & 14.12 & 19.5 & 200N & 56.72 & \$148 \textendash \ \$496 \\
ProxylessNAS + AMC - 8bit & 73.3 & 9.77 & 10.53 & 15.0 & 204N & 57.85 & \$151 \textendash \ \$506 \\
\hline
MobileNetV2 + HAQ & 71.9 & 8.93 & 11.82 & - & 96N & 27.23 & \$71 \textendash \ \$238 \\
ProxylessNAS + AMC + HAQ & 71.8 & 8.45 & 8.84 & - & 300N & 85.08 & \$222 \textendash \ \$744 \\
DNAS~\cite{wu2018mixed} & 74.0 & - & - & 57.3 & 40N & 11.34 & \$30 \textendash \ \$99 \\
Single Path One-Shot~\cite{guo2019single} & 74.6 & - & - & 51.9 & 288 + 24N & 6.81 & \$18 \textendash \ \$60 \\
\hline 
Ours-A (w/o transfer) & 72.1 & 8.85 & 11.79 & 13.2 & 2400 + 0.5N & \textbf{0.14} & \textbf{\$0.4 \textendash \ \$1.2} \\
Ours-B (w/ transfer) & \textbf{74.1} & 8.40 & 12.18 & 16.5 & 2400 + 0.5N & \textbf{0.14} & \textbf{\$0.4 \textendash \ \$1.2} \\
Ours-C (w/ transfer) & \textbf{75.1} & 12.17 & 14.14 & 23.6 & 2400 + 0.5N & \textbf{0.14} & \textbf{\$0.4 \textendash \ \$1.2} \\
\hline
\end{tabular}
\label{tab:comparison_with_nas}
\end{table*}
\begin{figure*}[t]
    \centering
    \includegraphics[width=0.85\linewidth]{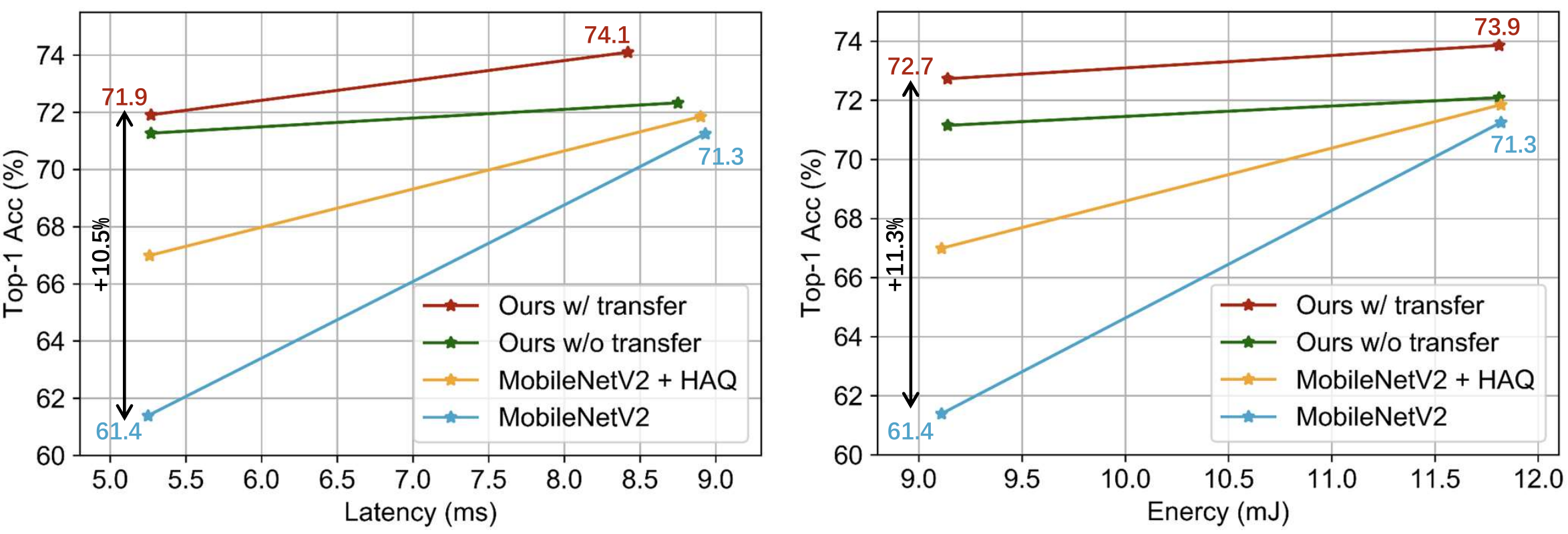}
    \caption{Comparison with mixed-precision models searched by HAQ~\cite{wang2019haq} under latency/energy constraints. The baselines are 4-bit and 6-bit fixed precision, respectively. When the constraint is strict, our model can outperform fixed precision model by more than 10\% accuracy, and 5\% compared with HAQ. Such performance boost may benefit from the dynamic architecture search space rather than fixed one as MobileNetV2.}
\label{fig:latency_and_energy}
\end{figure*}
\myparagraph{Quantization.}
We follow the implementation in~\cite{wang2019haq} to do quantization. Specifically, we quantize the weights and activations with the specific quantization policies. For each layer with weights $w$ with quantization bit $b$, we linearly quantize it to $[-v, v]$, the quantized weight is:
\begin{align}
    w' = \max(0,\min(2v, round(\frac{2w}{2^{b}-1})\cdot v)) - v
\end{align}
We set choose different $v$ for each layer that minimize the KL-divergence $\mathcal{D}(w || w')$ between origin weights $w$ and quantized weights $w'$.
For activation weights, we quantize it to $[0, v]$ since the value is non-negative after ReLU6 layer.

\section{Experiments}
\begin{figure*}[t]
    \centering
    \includegraphics[width=0.85\linewidth]{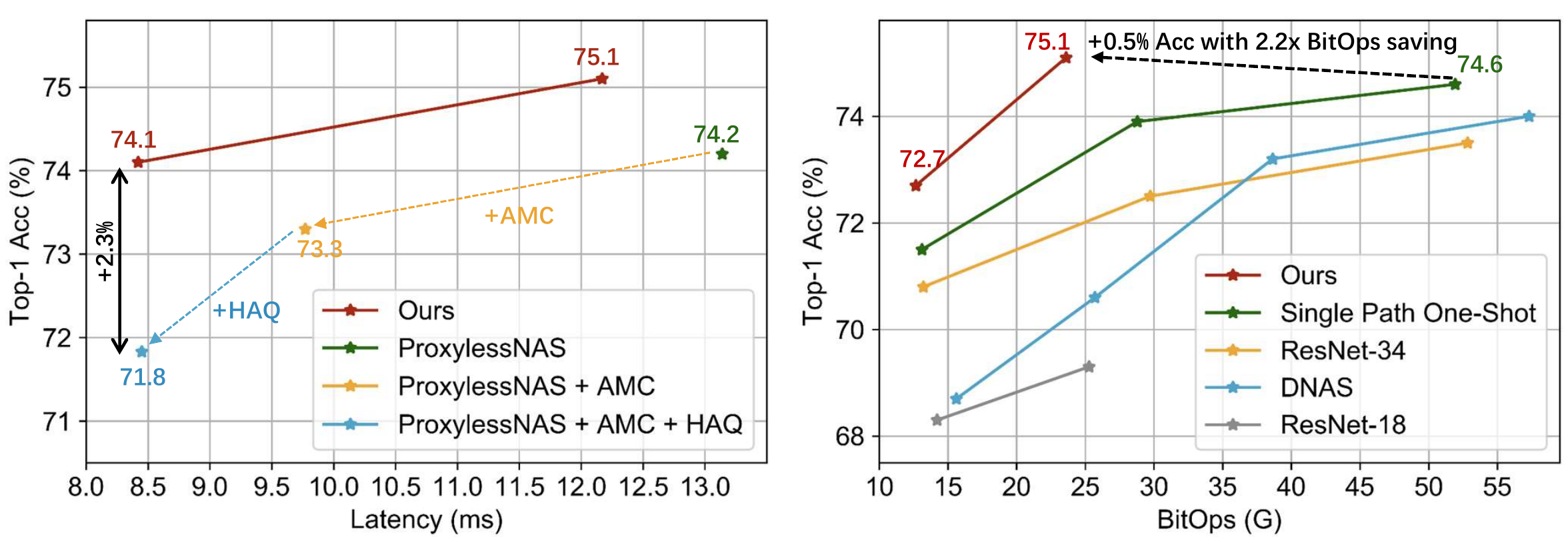}
    \caption{Left: Comparison with \textit{sequentially designed} mixed-precision models searched by ProxylessNAS, AMC and HAQ~\cite{cai2019proxylessnas,he2018amc,wang2019haq} under latency constraints. Our joint designed model achieves better accuracy than sequentially designed models. Right: Comparison with quantized model under BitOps constraint. The ResNet-34 baselines are 2/3/4 bit weight and activation. Our model achieves 0.5\% accuracy boost (from 74.6\% to 75.1\%) compared with models searched by single path one-shot while occupying half of BitOps. 
    }   
    \label{fig:compare_with_seq}
    \label{fig:bitops} 
\end{figure*}



\begin{figure*}[t]
    \vspace{-10pt}
    \centering
    \includegraphics[width=0.85\linewidth]{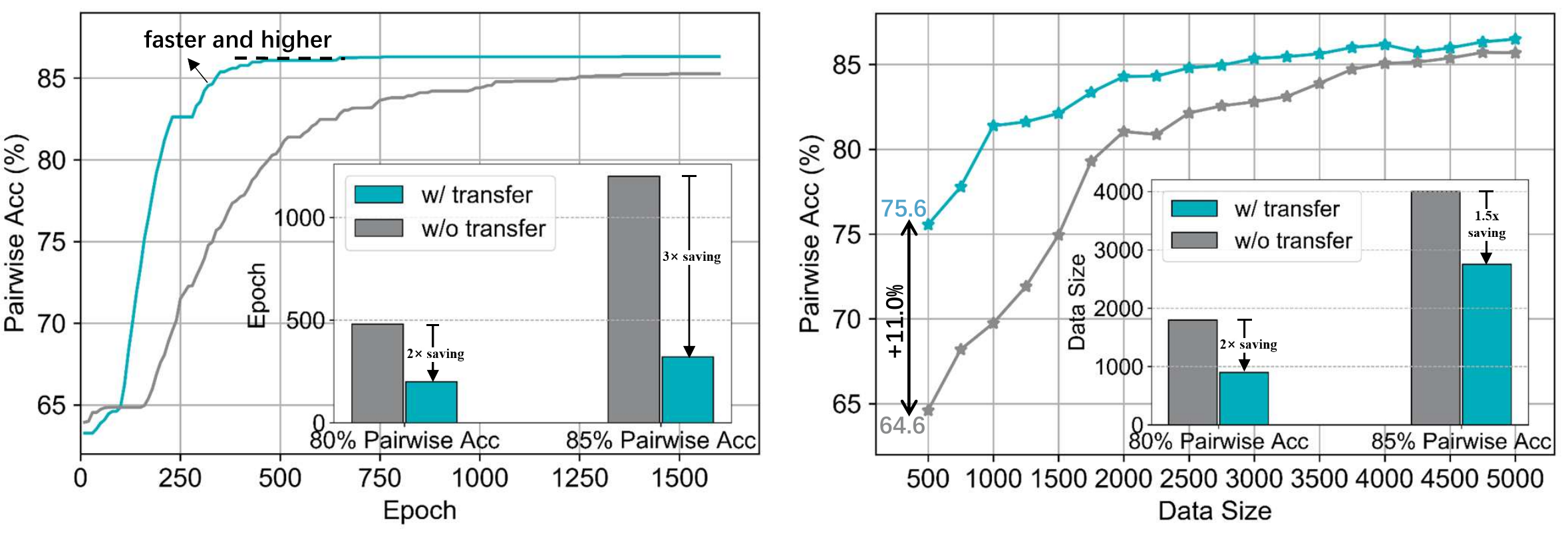}
    \caption{Illustration of the performance w/ or w/o predictor-transfer technique. Pairwise accuracy is a metric that measures the relative relationship between every two architectures. Left graph shows that the quantization-aware predictor could attain a faster and higher convergence with transferring. Right graph shows that when data is limited, predictor-transfer technique could largely improve the pairwise accuracy (from 64.6\% to 75.6\%). Using predictor-transfer technique, we can achieve 85\% pairwise accuracy using less than 3k data points, while at least 4k data will be required without this technique.}    \vspace{-10pt}
\label{fig:transfer_comp}
\end{figure*}
To verify the effectiveness of our methods, we conduct experiments that cover two of the most important constraints for on-device deployment: \textit{latency} and \textit{energy consumption} in comparison with some state-of-the-art models using neural architecture search. Besides, we compare BitOps with some multi-stage optimized models.
\myparagraph{Dataset, Models and Hardware Platform.}
The experiments are conducted on ImageNet dataset. We compare the performance of our joint designed models with mixed-precision models searched by~\cite{wang2019haq, he2018amc, cai2019proxylessnas} and some SOTA fixed precision 8-bit models. The platform we used to measure the resource consumption for the mixed-precision model is BitFusion~\cite{sharma2018bitfusion}, which is a state-of-the-art spatial ASIC design for neural network accelerator. It employs a 2D systolic array of Fusion Units which spatially sum the shifted partial products of two-bit elements from weights and activations.
\subsection{Comparison with SOTA Efficient Models}
Table~\ref{tab:comparison_with_nas} presents the results for different efficiency constraints. As one can see, our model can consistently outperform state-of-the-art models with either fixed or mixed-precision. Specifically, our small model (Ours-B) can have 2.2\% accuracy boost than mixed-precision MobileNetV2 search by HAQ (from 71.9\% to 74.1\%); our large model (Ours-C) attains better accuracy (from 74.6\% to 75.1\%) while only requiring half of BitOps. When applied with transfer technology, it does help for the model to get better performance (from 72.1\% to 74.1\%). It is also notable that the marginal cost for cloud computer and CO$_2$ emission is two orders of magnitudes smaller than other works.
\subsection{Effectiveness of Joint Design}
\vspace{10pt}
\myparagraph{Comparison with MobileNetV2+HAQ.}
Figure~\ref{fig:latency_and_energy} show the results on the BitFusion platform under different latency constraints and energy constraints.
Our jointly designed models consistently outperform both mixed-precision and fixed precision SOTA models under certain constraints. It is notable when the constraint is tight, our models have significant improvement compared with state-of-the-art mixed-precision models. Specifically, with similar efficiency constraints, we improve the ImageNet top1 accuracy from the MobileNetV2 baseline 61.4\% to 71.9\% (+10.5\%) and 72.7\% (+11.3\%) for latency and energy constraints, respectively. Moreover, we show some models searched by our quantization-aware predictor without predictor-transfer technique. With this technique applied, the accuracy can consistently have an improvement, since the non-transferred predictor might lose some mutual information between architecture and quantization policy.
\myparagraph{Comparison with Multi-Stage Optimized Model.}
Figure~\ref{fig:compare_with_seq} 
compares the multi-stage optimization with our joint optimization results.
As one can see, under the same latency/energy constraint, our model can attain better accuracy than the multi-stage optimized model (74.1\% vs 71.8\%). This is reasonable since the per-stage optimization might not find the global optimal model as the joint design does.
\myparagraph{Comparison under Limited BitOps.}

Figure \ref{fig:bitops} reports the results with limited BitOps budget. As one can see, under a tight BitOps constraint, our model improves over 2\% accuracy (from 71.5\% to 73.9\%) compared with the searched model using~\cite{guo2019single}. 
Moreover, our models achieve even higher accuracy (75.1\%) as ResNet34 8-bit model (75.0\%) while saving 8 $\times$ BitOps.

\subsection{Effectiveness of Predictor-Transfer}
Figure~\ref{fig:transfer_comp} shows the performance of our predictor-transfer technique compared with training from scratch. For each setting, we train the predictor to convergence and evaluate the pairwise accuracy (\ie the proportion that predictor correctly identifies which is better between two randomly selected candidates from a held-out dataset), which is a measurement for the predictor's performance. We use the same test set with 2000 $\langle$NN architecture, ImageNet accuracy$\rangle$ pairs that are generated by randomly choosing network architecture and quantization policy. Typically, for training with $N$ data points, the number of two kinds of data as mentioned in Sec. \ref{sec:detail} is equal, i.e., $N/2$. As shown, the transferred predictor has a higher and faster pairwise accuracy convergence. Also, when the data is very limited, our method can achieve more than 10\% pairwise accuracy over scratch training.

\section{Conclusion}
We propose \textit{\modelshort}, a joint design method for architecting mixed-precision model. Unlike former works that decouple into separated stages, we directly search for the optimal mixed-precision architecture without multi-stage optimization. We use a predictor-base method that can have no extra evaluation for the target dataset, which greatly saves GPU hours for searching under an upcoming scenario, thus reducing marginally CO$_2$ emission and cloud compute cost. To tackle the problem for the high expense of data collection, we propose a predictor-transfer technique to make up for the limitation of data. Comparisons with state-of-the-art models show the necessity of joint optimization and the prosperity of our joint design method.

\subsubsection*{Acknowledgments}
We thank  NSF Career Award \#1943349, MIT-IBM Watson AI Lab, Samsung, SONY, SRC, AWS Machine Learning Research Award  for supporting this research. We thank Hanrui Wang and Yujun Lin for their kindly help to this paper.

{\small
\bibliographystyle{ieee_fullname}
\bibliography{egbib}
}

\end{document}


\begin{figure*}[t]
\vspace{-20pt}
    \includegraphics[width=1\linewidth]{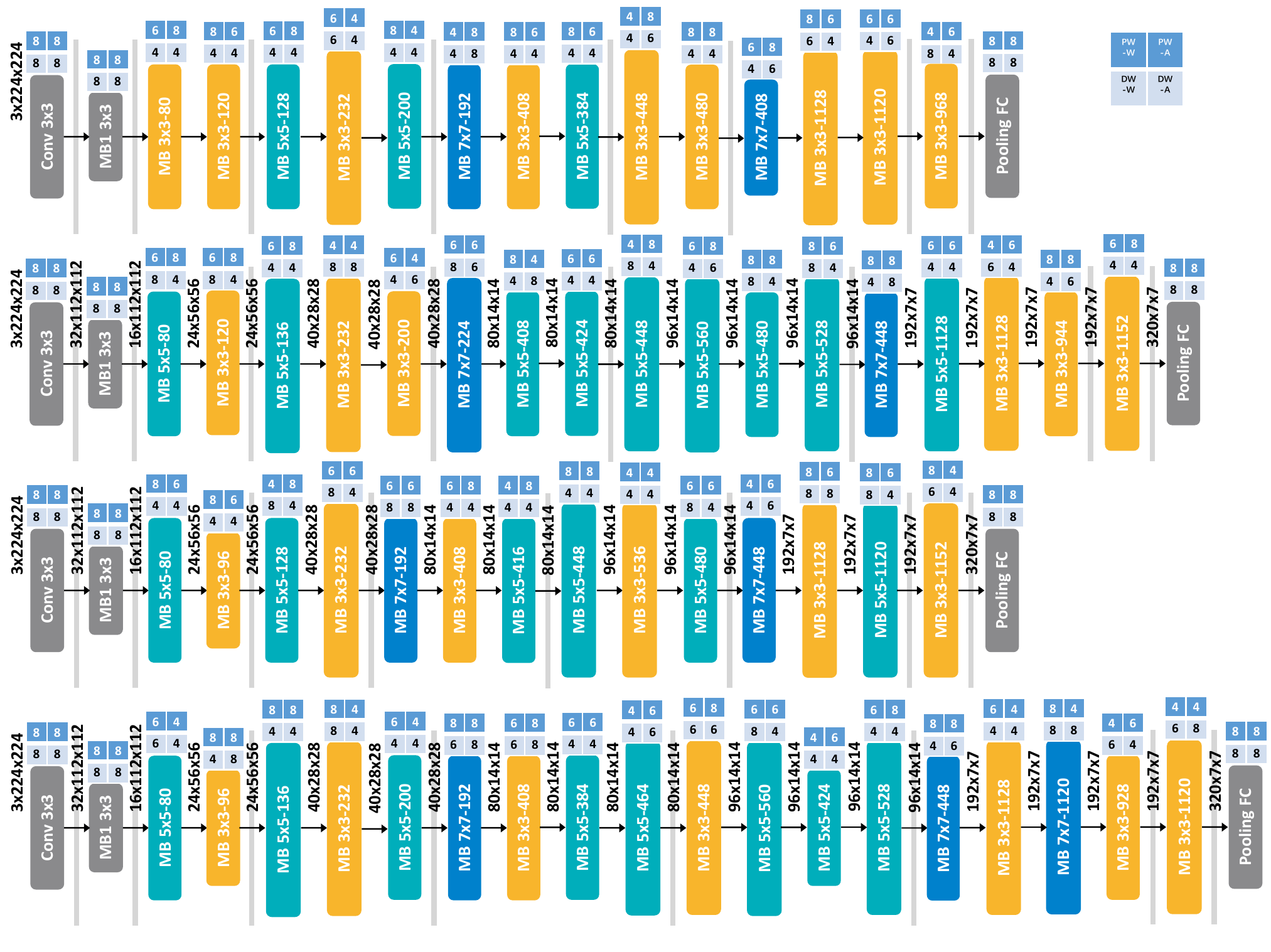}
    \caption{Illustration of the network architecture searched by APQ. PW/DW-W/A denotes the bit width for pointwise/depthwise convolution layer's weight/activation. The models are searched under 4-bit MobileNetV2 latency constraint, 6-bit MobileNetV2 latency constraint, 4-bit MobileNetV2 energy constraint, 6-bit MobileNetV2 energy constraint, respectively. One can see that shallow layer models are searched under a tight resource budget while deeper models are searched when the budget is loose, which is in line with intuition. The implementation for APQ can be found at \url{https://drive.google.com/drive/folders/1wbPQSPC-rfivH8f0c80s8hKBwemq8c9N?usp=sharing}
    }

\label{fig:network_arch}
\end{figure*}